\documentclass[11pt]{article}

\usepackage[final]{acl}

\usepackage{times}
\usepackage{latexsym}

\usepackage[T1]{fontenc}

\usepackage[utf8]{inputenc}

\usepackage{microtype}

\usepackage{inconsolata}

\usepackage{graphicx}
\usepackage[utf8]{inputenc} 
\usepackage[T1]{fontenc}    

\usepackage{microtype}
\usepackage{hyperref}
\usepackage{url}
\usepackage{booktabs}
\usepackage{lineno}
\usepackage{amsmath}
\usepackage{amssymb}
\usepackage{mathtools}
\usepackage{amsthm}
\usepackage{array}
\usepackage{enumitem}
\usepackage{longtable}
\usepackage{xparse}
\usepackage{ragged2e}
\usepackage{microtype}
\usepackage{xcolor}
\usepackage{graphicx}
\usepackage{colortbl}
\usepackage{pifont}
\usepackage{enumitem}
\usepackage{wrapfig}
\usepackage[utf8]{inputenc}
\usepackage[most]{tcolorbox}
\tcbuselibrary{listings, skins, breakable, theorems}
\tcbuselibrary{breakable}
\definecolor{royalblue}{RGB}{0, 70, 180}
\definecolor{olivegreen}{RGB}{0, 110, 0}
\usepackage{algorithm}
\usepackage{algpseudocode}
\usepackage{lipsum}
\usepackage{multirow}
\usepackage{caption}
\usepackage{adjustbox}
\usepackage{float}
\usepackage{stfloats} 

\newcommand{\red}[1]{\textcolor{red}{#1}}

\definecolor{darkblue}{rgb}{0, 0, 0.5}
\hypersetup{colorlinks=true, citecolor=darkblue, linkcolor=darkblue, urlcolor=darkblue}

\title{Two Heads are Better Than One:\\ Test-time Scaling of Multi-agent Collaborative Reasoning}

\author{
\mdseries
Can Jin\textsuperscript{1*},
Hongwu Peng\textsuperscript{2},
Qixin Zhang\textsuperscript{3},
Yang Zhou\textsuperscript{1},
Yujin Tang\textsuperscript{4} \\
Tong Che\textsuperscript{5\textdagger},
Dimitris N. Metaxas\textsuperscript{1\textdagger}
\\
\textsuperscript{1}Rutgers University \quad
\textsuperscript{2}University of Connecticut \quad
\textsuperscript{3}Nanyang Technological University \\
\textsuperscript{4}Sakana AI \quad
\textsuperscript{5}NVIDIA Research
\\
\textsuperscript{*}Correspondence to: \texttt{can.jin@rutgers.edu} \quad
\textsuperscript{\textdagger}Equal Advising
}

\begin{document}
\maketitle
\begin{abstract}
Test-Time Scaling has emerged as a powerful method to extend the reasoning capabilities of Large Language Models. However, single-agent TTS faces significant scalability bottlenecks, as excessively long reasoning traces lead to increased inference costs and stability issues caused by context management failures. To address these limitations, we propose leveraging Multi-Agent Systems as a structural upgrade to standard TTS. By decomposing monolithic reasoning chains into distinct, manageable contexts across multiple agents, MAS offers a more robust framework for scaling reasoning. We validate this approach by introducing M500, a dataset comprising 500 high-quality multi-agent, multi-turn collaborative reasoning traces generated via DeepSeek-R1. Through Supervised Fine-Tuning on M500, we enable open-source models to internalize collaborative reasoning patterns and show improved TTS performance in MAS. Furthermore, we propose an adaptive scaling strategy incorporating a ``CEO'' agent to dynamically guide the reasoning process and optimize collaboration depth. Extensive experiments within the AgentVerse framework demonstrate that our fine-tuned models, Qwen2.5-32B-MAS and Phi4-14B-MAS, significantly outperform their base counterparts. Codes are available at \url{https://github.com/jincan333/MAS-TTS.}
\end{abstract}
\section{Introduction}
The advancement of Large Language Models (LLMs)~\citep{gpt4,gpt4o,deepseek_v3,qwen2,abdin2024phi,jin2024learning} has significantly extended the capabilities of artificial intelligence. Recently, Test-Time Scaling (TTS) has emerged as an effective approach to further enhance LLM performance, particularly for complex mathematical reasoning tasks~\citep{r1,snell2025scaling,tts,s1,rebase,jin2026weak}. Techniques such as Monte Carlo Tree Search~\citep{mcts,zhang2024o1codero1replicationcoding}, large-scale reinforcement learning~\citep{r1,o1,o1rl}, and supervised fine-tuning (SFT) on detailed reasoning chains~\citep{s1,limo} have been widely applied to extend the inference process. However, extending the reasoning trace for a single agent introduces significant challenges. As the reasoning chain grows excessively long, the inference cost increases substantially\citep{dao2022flashattention}, and the model becomes prone to divergence and hallucination due to the difficulty of managing a massive context window effectively\citep{liu2024lost, dziri2023faith, turpin2023language}.

To address the limitations of single-agent TTS, we propose leveraging Multi-Agent Systems (MAS) as a structural extension to standard TTS. The core advantage of MAS in this context is the ability to decompose a potentially unstable, monolithic reasoning trace into distinct segments managed by different roles~\citep{feng2025one,camel}. By assigning specific contexts and sub-tasks to different agents, MAS can maintain coherence over long-form reasoning processes that would otherwise overwhelm a single agent \citep{agentverse,chatdev,agent_laboratory,ai_scientist}. Consequently, MAS serves as a more robust framework for TTS, allowing for effective scaling of reasoning capabilities without the context management failures observed in single-agent settings.

Building on the concept of Simple-Scaling~\citep{s1}, we hypothesize that the collaborative reasoning capabilities required for this enhanced TTS can be established via SFT on multi-agent Chain-of-Thought (CoT) interaction data. To investigate this, we first curate a high-quality multi-agent collaborative reasoning dataset by solving a diverse set of challenging problems using an open-source MAS. Each instance in the dataset comprises the interactive sequence of multi-agent, multi-turn data generated by the MAS to solve a specific problem. To ensure dataset quality and support long-form CoT, we filter low-quality examples and use DeepSeek-R1~\citep{r1} to generate robust reasoning traces for every agent. We designate this dataset as M500, which contains 500 diverse and detailed multi-agent collaborative reasoning traces. We then perform SFT on M500 using standard open-source, instruction-tuned LLMs. To further optimize adaptive reasoning and scaling in the MAS, we introduce a ``\textbf{CEO}'' agent powered by the SFT model, inspired by the observation that leaderless groups in human societies often lack effectiveness and coherent direction~\citep{cruz1999impact,hackman2002leading}. This agent dynamically guides discussions, effectively managing collaborative efforts and reasoning depth to improve the overall performance of the system.

We conduct extensive experiments to validate our approach by fine-tuning Qwen2.5-32B-Instruct~\citep{qwen25} and Phi4-14B~\citep{abdin2024phi} on M500, yielding the models Qwen2.5-32B-MAS and Phi4-14B-MAS. We then integrate these models into the AgentVerse~\citep{agentverse} multi-agent framework. Testing across various task categories—including general understanding, mathematical reasoning, and coding—demonstrates that our Qwen2.5-32B-MAS and Phi4-14B-MAS significantly surpass the baseline Qwen2.5-32B-Instruct and Phi4-14B within the MAS. For example, our method achieves a 12\% improvement on GPQA-Diamond~\citep{gpqa}, a 41\% improvement on AIME2024~\citep{aime2024}, and a 10\% improvement on MBPP-Sanitized~\citep{mbpp}, while matching the performance of DeepSeek-R1 on MATH-500 and MBPP-Sanitized.

In summary, our contributions are: \ding{182} We develop M500, a comprehensive multi-agent, multi-turn collaborative reasoning dataset created via an automatic generation pipeline; \ding{183} We demonstrate the collaborative reasoning capability of LLMs in MAS can be effectively improved by SFT on multi-agent, multi-turn reasoning datasets, serving as an effective implementation of MAS-TTS; \ding{184} We propose an adaptive scaling strategy that incorporates a CEO agent powered by the obtained LLMs to dynamically guide and scale multi-agent collaborative reasoning; and \ding{185} We show through extensive experiments that our method significantly outperforms baseline models and achieves performance comparable to DeepSeek-R1 on specific tasks.
\section{Related Works}
\subsection{Test-time Scaling}
A wide range of methods have been developed to improve reasoning in LLMs by leveraging Test-Time Scaling (TTS). Recent work explores techniques including hierarchical hypothesis search, which enables inductive reasoning through structured exploration \citep{wang2023hypothesis,li2026latent}. They also examine tool augmented inference, which enhances downstream performance by allowing models to interact with external environments \citep{gao2023pal, qintoolllm}. Other approaches focus on internal mechanisms, such as learning thought tokens in an unsupervised manner \citep{zelikman2024quiet, goyal2023think}, allowing models to better utilize extended reasoning sequences. Among the most studied scaling paradigms are parallel and sequential TTS approaches. Parallel methods generate multiple solution candidates independently and select the best one using a scoring criterion, such as majority voting or outcome-based reward models \citep{brown2024large, irvine2023rewarding, snell2025scaling}. In contrast, sequential methods condition each new attempt on the previous ones, allowing iterative refinement based on prior outputs \citep{s1,snell2025scaling, hou2025advancing, lee2025evolving}. Bridging these strategies, tree-based techniques such as Monte Carlo Tree Search (MCTS) \citep{zhang2023planning, zhou2023language, zhang2024llama, rebase} and guided beam search \citep{xie2023self} enable structured exploration through branching and evaluation. Central to many of these methods are reward models, which provide feedback signals during generation. These can be categorized as outcome reward models, which evaluate entire solutions \citep{xin2024deepseek, ankner2024critique}, or process reward models, which assess intermediate reasoning steps \citep{lightman2023let, wang2023math, rebase}, guiding the model toward more effective reasoning paths.

\subsection{LLM Agents}
Recent work has extended the capabilities of LLMs beyond standalone reasoning and understanding, enabling them to operate as multi-agents that can interact with environments, tools, and other agents to perform complex tasks \citep{wu2023autogen, camel, agentverse, chatdev, motwani2024malt, jin2025rankflow, jin2025apeer}. These multi-agent systems (MAS) integrate various techniques, including CoT prompting \citep{wei2022chain}, iterative refinement \citep{shinn2023reflexion}, self-improvement \citep{huang2023large}, and external tool usage \citep{hao2023toolkengpt, qintoolllm, schick2023toolformer}, to support multi-step decision-making and long-horizon planning. They have been applied successfully in domains such as mathematical reasoning \citep{agentverse}, software engineering \citep{chatdev, jimenez2023swe, wang2024opendevin}, and scientific discovery \citep{ai_scientist, agent_laboratory}. Agent frameworks typically structure the interaction with LLMs using techniques such as few-shot prompting \citep{brown2020language} and guided reasoning \citep{wei2022chain,shinn2023reflexion}, relying on the model’s in-context learning capabilities \citep{olsson2022context}.
\section{Methodology}
We first describe the automatic generation of high-quality multi-agent collaborative reasoning data. Next, we improve the collaborative reasoning capabilities of LLMs by performing SFT on the generated data. Finally, we introduce a CEO agent into the MAS to further enable adaptive scaling by directing collaboration and adjusting resource allocation. Our method overview is in Figure \ref{fig:ceo}.

\begin{figure}[t]
\centering
\begin{tcolorbox}[
  enhanced,
  title=Multi-agent Collaborative Reasoning Trace,
  colback=white,              
  colframe=black,             
  colbacktitle=gray!5,       
  coltitle=black,             
  fonttitle=\bfseries,        
  boxrule=0.5pt,              
  sharp corners,
  left=2pt, right=2pt, top=1pt, bottom=1pt
]
\scriptsize

\begin{tcolorbox}[
  enhanced,
  colback=white,
  colframe=royalblue,
  boxrule=0.5pt,
  sharp corners,
  left=1pt, right=1pt, top=1pt, bottom=1pt
]
\textcolor{royalblue}{\textbf{Expert Recruiter}} \\
\textbf{Prompt:} You are leading a group of experts to solve: ... Find the greatest integer that does not exceed $\dfrac{x^2}{100}$. Recruit two experts with different specialties to best solve the problem... \\
\textbf{Response:}
\textless think\textgreater\space ... I need to identify experts best suited for this geometry task... \textless/think\textgreater 1. A geometric analysis expert. 2. An algebraic modeling expert.
\end{tcolorbox}

\begin{tcolorbox}[
  enhanced,
  colback=white,
  colframe=olivegreen,
  boxrule=0.5pt,
  sharp corners,
  left=1pt, right=1pt, top=1pt, bottom=1pt
]
\textcolor{olivegreen}{\textbf{Problem Solver 1 (Geometric Expert)}} \\
\textbf{Prompt:} ... Solve the following accurately: One base of a trapezoid is $100$ units longer than the other... Find the greatest integer that does not exceed $x^2/100$... \\
\textbf{Response:}
\textless think\textgreater\space ... Let’s understand the structure: the segment splits the area in a 2:3 ratio. We need the length $x$ such that both halves have equal area... \textless/think\textgreater ...Using RMS to find $x$, we get:
{\tiny
\vspace{-2mm}
\[
\frac{x^2}{100} = \frac{18125}{100} = 181.25 \Rightarrow \boxed{181}
\]
}
\end{tcolorbox}

\begin{tcolorbox}[
  enhanced,
  colback=white,
  colframe=orange,
  boxrule=0.5pt,
  sharp corners,
  left=1pt, right=1pt, top=1pt, bottom=1pt
]
\textcolor{orange}{\textbf{Problem Solver 2 (Algebraic Expert)}} \\
\textbf{Prompt:} ... Given the prior solution, critique its correctness. Do not solve again. If you agree with the final answer, append \texttt{[Agree]}... \\
\textbf{Response:}
\textless think\textgreater\space ... Reviewing the solution: the answer is correct. However, explanation for using RMS could be clearer... \textless/think\textgreater ... The reasoning is sound and the steps are accurate. \texttt{[Agree]}
\end{tcolorbox}

\begin{tcolorbox}[
  enhanced,
  colback=white,
  colframe=red,
  boxrule=0.5pt,
  sharp corners,
  left=1pt, right=1pt, top=1pt, bottom=1pt
]
\textcolor{red}{\textbf{Evaluator}} \\
\textbf{Prompt:} You are an experienced science teacher. Evaluate the final solution from both solvers for correctness... \\
\textbf{Response:}
\textless think\textgreater\space ... Let me verify each step. Yes, the use of RMS and the calculation are both appropriate... \textless/think\textgreater ... \textbf{Correctness:} 1 ...
\end{tcolorbox}
\end{tcolorbox}
\caption{An illustration of a multi-agent collaborative reasoning data sample from M500.}
\label{fig:data_sample}
\end{figure}

\subsection{Automatic Generation of Multi-Agent Collaborative Reasoning Data}\label{subsec_generate_reasoning_trace}

\paragraph{Question Sampling Based on Difficulty, Diversity, and Interdisciplinarity.}
When selecting questions for our multi-agent collaborative reasoning dataset, we consider three main aspects: \ding{182} Difficulty, \ding{183} Diversity, and \ding{184} Interdisciplinarity. We begin with the full Simple-Scaling dataset \citep{s1}, which includes diverse questions sourced from historical AIME problems, OlympicArena \citep{olympicarena}, and AGIEval \citep{agieval}, among others. These questions span multiple domains, including physics, geometry, number theory, biology, and astronomy. To promote difficulty and interdisciplinarity, we use DeepSeek-R1 \citep{r1} to determine whether solving each question requires interdisciplinary knowledge, excluding those that DeepSeek-R1 answers using fewer than 1024 tokens. Questions selected through this process are generally challenging and demand knowledge integration from multiple disciplines. For example, solving a complex mathematics problem might benefit from collaboration between algebra and geometry specialists, whereas an advanced astronomy question may require input from astronomers, physicists, and mathematicians.

\paragraph{Generating Multi-Agent Collaborative Reasoning Traces.}
To generate collaborative reasoning traces, we employ the open-source MAS frameworks and reasoning models, specifically AgentVerse \citep{agentverse} and DeepSeek-R1 \citep{r1}, to process previously selected questions. This MAS framework involves multiple roles: Expert Recruiter (e.g., Human Resources), Problem Solvers (e.g., Scientists and Engineers), Executor (e.g., Quality Assurance Engineers), and Evaluator (e.g., Product Manager). The agents collaborate in the following manner: the Expert Recruiter identifies and assigns suitable experts for the task, while keeping the \textbf{total agent number} fixed and independent of question complexity or available information. These recruited experts function as Problem Solvers, proposing and iteratively refining solutions through multi-turn discussions with a predetermined \textbf{number of critic iterations}. Once a consensus is reached, or if the maximum number of critic iterations is reached, the resulting solution is passed to the Executor, who runs the necessary code or invokes external tools if required. The Evaluator then reviews both the solution and the results produced by the Executor, providing feedback for further refinement. This iterative process continues until either the solution is deemed correct or the predefined \textbf{maximum total number of iterations} is reached. Each MAS role produces distinct inputs and outputs, generating reasoning traces from its expert viewpoint. We collect all input-output pairs from each agent and compile the entire collaborative reasoning history into a comprehensive trace for each question.

\vspace{-2mm}
\begin{figure*}[ht]
    \centering
    \includegraphics[width=0.9\linewidth]{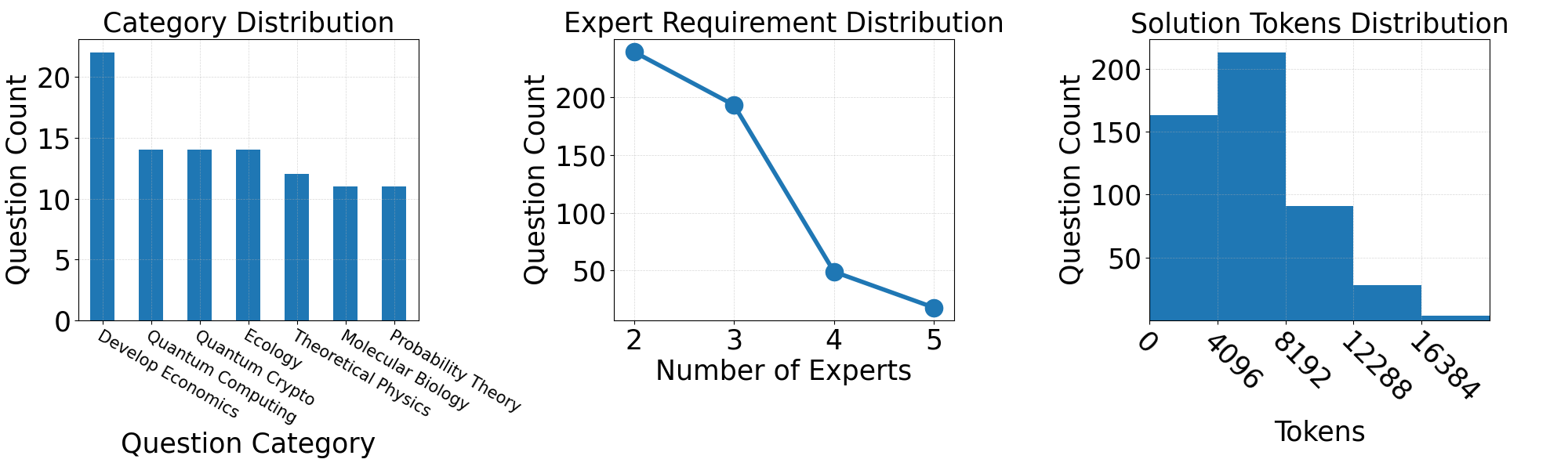}
    \caption{Distributions of key statistics in M500: question category (filtered with count $>10$), predicted number of experts required for solving each problem, and solution token usage.}
    \label{fig:data_analyze}
\end{figure*}

\paragraph{Data Filtering.}
To ensure high-quality collaborative reasoning traces, we filter data based on three criteria: \ding{182} \textbf{Consensus Reached:} Traces where Problem Solvers fail to reach consensus within the maximum allowed critic iterations are discarded. This criterion ensures effective discussion convergence and minimizes unresolved reasoning. \ding{183} \textbf{Format Compliance:} Samples that deviate from the required format are excluded. Specifically, each agent's reasoning must be enclosed within \texttt{<think>} and \texttt{</think>} tags, and the final answers must be encapsulated within \texttt{boxed\{\}}. This maintains output consistency and facilitates automated parsing and answer extraction. \ding{184} \textbf{Correctness:} We extract the final answer from each collaborative trace and compare it against the ground-truth solution, discarding any traces with incorrect answers.

Through this method, we iteratively sample 500 questions along with their corresponding multi-agent collaborative reasoning traces, forming the \textbf{M500} dataset. This dataset contains 500 challenging and diverse questions requiring interdisciplinary collaboration, accompanied by comprehensive and high-quality reasoning traces that document the complete multi-agent problem-solving process. The pseudocode detailing this sampling procedure is provided in Algorithm \ref{algo:data_generation} in the Appendix.

Figure \ref{fig:data_sample} shows an illustrative example from M500, with the complete data sample presented in Figure \ref{fig:data_sample_complete} in the Appendix. The collaborative reasoning trace for this example includes input-output interactions among four agents: Expert Recruiter, Geometry Expert, Algebra Expert, and Evaluator. The example question is sufficiently challenging (requiring 5695 tokens), achieves consensus among agents, complies with the required format, and produces a correct solution. Additionally, the distribution of question categories in the M500 dataset, predicted expert counts, and solution token usage is illustrated in Figure \ref{fig:data_analyze}. We observe significant diversity in the dataset across fields such as economics, physics, biology, and mathematics. Most questions are predicted to be solved by two experts and require fewer than 8192 tokens to solve.

\subsection{Enhancing LLM Collaborative Reasoning through Supervised Fine-Tuning}
Inspired by Simple-Scaling \citep{s1}, which shows that long-form CoT reasoning capabilities in LLMs can be developed through SFT on detailed reasoning traces, we apply SFT to an LLM $f$ using the M500 dataset. The goal is to enable $f$ to produce long-form CoT reasoning that supports the collaboration in a MAS. Specifically, the SFT objective is to minimize the following loss:
\[
\textstyle \mathcal{L}_{\text{SFT}} = \mathbb{E}_{(\mathbf{x}, \mathbf{y}) \in \text{M500}} [ -\frac{1}{|\mathbf{y}|} \sum_{t=1}^{|\mathbf{y}|} \log P_f(\mathbf{y}_t \mid \mathbf{x}, \mathbf{y}_{<t}) ]
\]
where \( P_f(\mathbf{y}_t \mid \mathbf{x}, \mathbf{y}_{<t}) \) denotes the probability the model \( f \) assigns to token \( \mathbf{y}_t \) given input \( \mathbf{x} \) and previous tokens \( \mathbf{y}_{<t} \).

For each question $q$ in the M500 dataset, we collect a series of input-output pairs $\{(\mathbf{x}_i, \mathbf{y}_i)\}_{i=1}^{n}$ corresponding to the reasoning traces from all participating agents. During training, we ensure all reasoning traces for $q$, $\{(\mathbf{x}_i, \mathbf{y}_i)\}_{i=1}^{n}$, are grouped within the same batch and ordered according to the original generation sequence in the MAS. This approach helps the model learn collaborative reasoning in a coherent and temporally consistent manner.

\vspace{-2mm}
\begin{figure*}[ht]
    \centering
    \includegraphics[width=0.95\linewidth]{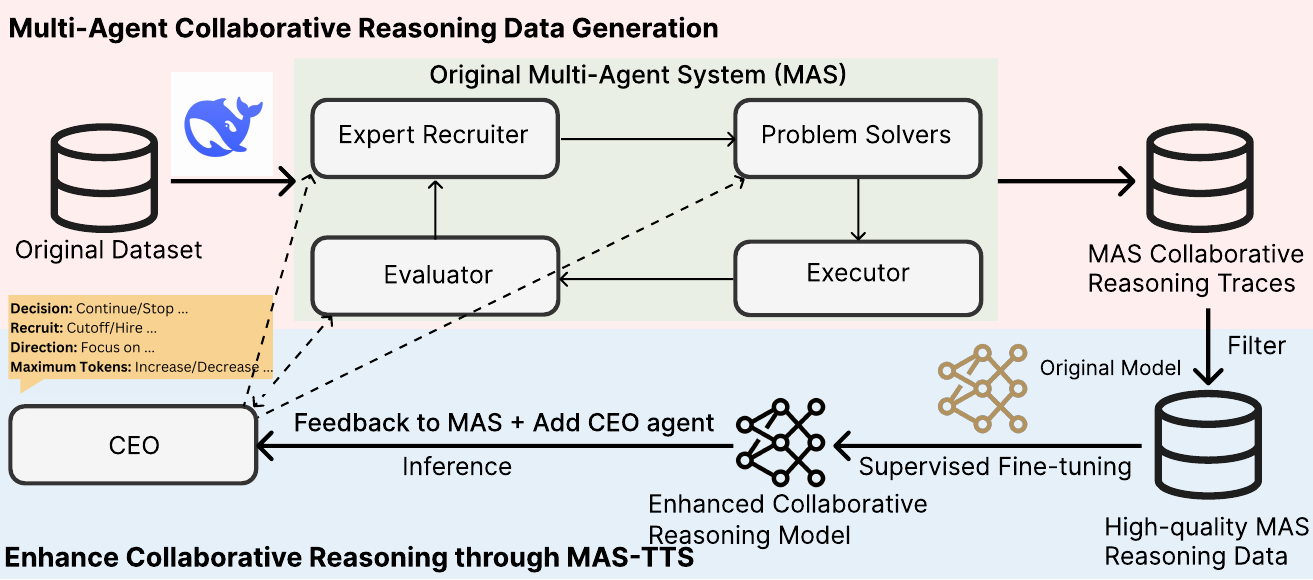}
    \caption{Overview of our method. First, we generate high-quality collaborative reasoning traces (M500) using the AgentVerse~\citep{agentverse} framework. We then perform SFT on the base model using M500 to improve its collaborative reasoning capabilities. Finally, we integrate the fine-tuned model back into the MAS and introduce a CEO agent to adaptively scale reasoning by dynamically adjusting collaboration intensity and reasoning depth.}
    \label{fig:ceo}
\end{figure*}

\subsection{Adaptive Test-time Scaling}

Recently, TTS has emerged as an effective method for enhancing the performance of LLMs. Models such as OpenAI's o-series and DeepSeek-R1 have shown considerable improvements by employing scaled reasoning during inference. However, the application of TTS within MAS remains relatively unexplored. Previous studies in single-agent scenarios indicate that the optimal TTS strategy depends on question difficulty \citep{tts,rebase}. In MAS, choosing an appropriate TTS strategy is even more critical due to the significantly higher computational and time costs involved in collaboration compared to single-agent.

To address this issue, we propose an adaptive TTS strategy for MAS by introducing a dedicated "\textbf{CEO}" agent, which dynamically manages collaboration and resource allocation based on the task's ongoing progress. As shown in Figure \ref{fig:ceo}, the CEO agent evaluates the question, current solution state, evaluation feedback, and available resources to determine whether a proposed solution should be accepted or needs further refinement. Additionally, this agent directs subsequent discussions, decides how many agents to involve, and sets appropriate reasoning depth (i.e., the token budget for each agent's response).

Unlike static MAS configurations, which have fixed numbers of agents, iteration limits, and reasoning depths, our adaptive approach allows the MAS to dynamically adjust its settings. This capability enables more effective scaling of collaborative reasoning by modifying agent participation, termination conditions, and reasoning depth according to the evolving requirements of the problem.

\section{Experiments}\label{sec_experiments}
To validate that our method effectively improves collaborative reasoning in MAS, we conduct experiments using state-of-the-art (SOTA) open-source and closed-source LLMs within the AgentVerse framework. We evaluate performance across general understanding, mathematical reasoning, and coding tasks. Additionally, we analyze the emergent behavior and scaling properties of the proposed method.

\subsection{Experimental Details}\label{sec:experimental_details}
\paragraph{LLMs.} We evaluate both reasoning and non-reasoning LLMs to assess the impact of collaborative reasoning in MAS. Our primary baselines are Qwen2.5-32B-Instruct~\citep{qwen25,yang2025qwen3} and Phi4-14B~\citep{abdin2024phi}. To isolate the benefits of MAS-specific SFT, we compare our models against single-agent SFT baselines: Qwen2.5-32B-SFT and Phi4-14B-SFT. These baselines are fine-tuned on the same problem set as M500 but utilize direct, single-agent solutions generated by DeepSeek-R1. Additionally, we include s1.1-32B, which is trained on questions from Simple-Scaling~\citep{s1} using DeepSeek-R1 generated responses. Our models, Qwen2.5-32B-MAS and Phi4-14B-MAS, along with their respective SFT baselines, are fine-tuned from Qwen2.5-32B-Instruct and Phi4-14B, respectively. We also benchmark against strong open-source models, DeepSeek-V3~\citep{deepseek_v3} and DeepSeek-R1~\citep{r1}, as well as closed-source models, including o3-mini (medium)~\citep{o3-mini} and GPT-4o (2024-08-06)~\citep{gpt4o}.

\vspace{-2mm}
\begin{table*}[t]
\centering
\resizebox{1\textwidth}{!}{%
\begin{tabular}{lccccccc}
\toprule
\multirow{2}{*}{\textbf{Model}} & \multicolumn{2}{c}{\textbf{General Understanding}} & \multicolumn{2}{c}{\textbf{Mathematical Reasoning}} & \multicolumn{2}{c}{\textbf{Coding}} \\
\cmidrule(lr){2-3} \cmidrule(lr){4-5} \cmidrule(lr){6-7}
 & GPQA & Commongen & AIME2024 & MATH-500 & HumanEval & MBPP-S & Avg. \\
\midrule
\multicolumn{8}{c}{\textit{API-accessed Models}} \\
\midrule
DeepSeek-V3           & 58.6 & 98.6 & 33.3 & 88.6 & 89.6 & 83.9 & 75.4 \\
GPT-4o                & 49.2 & 97.8 & 7.8  & 81.3 & 90.9 & 85.4 & 68.7 \\
DeepSeek-R1           & \textbf{75.5} & 97.2 & 78.9 & \textbf{96.2} & \textbf{98.2} & 91.7 & 89.6 \\
o3-mini               & 71.3 & \textbf{99.1} & \textbf{84.4} & 95.3 & 97.0 & \textbf{93.6} & \textbf{90.1} \\
\midrule
\multicolumn{8}{c}{\textit{Locally deployed models}} \\
\midrule
Phi4-14B & 53.5 & 95.4 & 13.3 & 80.6 & 83.2 & 79.1 & 67.5 \\
Phi4-14B-SFT & 56.3 & 95.2 & 50.0 & 84.3 & 85.4 & 81.8 & 75.5 \\
Qwen2.5-32B-Instruct & 50.2 & 96.7 & 21.1 & 84.4 & 89.0 & 80.2 & 70.3 \\
Qwen2.5-32B-SFT & 56.1 & 96.2 & 53.3 & 88.6 & 85.2 & 84.1 & 77.3 \\
s1.1-32B & 58.3 & 94.1 & 53.3 & 90.6 & 82.3 & 77.4 & 76.0 \\
\midrule
Phi4-14B-MAS (Ours)               & 58.1 & 95.8 & 56.7 & 88.2 & 85.8 & 83.2 & 78.0 \\
Phi4-14B-MAS w. CEO (Ours)         & 58.8 & 96.2 & 60.0 & 89.8 & 87.5 & 84.6 & 79.5 \\
Qwen2.5-32B-MAS (Ours)               & 61.1 & 96.9 & 60.0 & 95.1 & 92.8 & 89.1 & 82.5 \\
Qwen2.5-32B-MAS w. CEO (Ours)         & \textbf{62.1} & \textbf{97.4} & \textbf{62.2} & \textbf{95.8} & \textbf{93.9} & \textbf{90.5} & \textbf{83.7} \\
\bottomrule
\end{tabular}}
\caption{Performance comparison on general understanding, mathematical reasoning, and coding tasks using strong reasoning and non-reasoning models within the AgentVerse framework. Our method achieves substantial improvements over baselines on all tasks and attains performance comparable to o3-mini and DeepSeek-R1 on MATH-500 and MBPP-S, demonstrating its effectiveness in enhancing collaborative reasoning in MAS.
}
\label{tab:main_results}
\end{table*}

\paragraph{Tasks.} We conduct a comprehensive evaluation across three critical domains: \ding{182} \textbf{General Understanding}: We employ GPQA-Diamond (GPQA)~\citep{gpqa} to assess general knowledge and Commongen-Challenge (Commongen)~\citep{self-refine} to evaluate writing capability. GPQA comprises 198 PhD-level science questions in Biology, Chemistry, and Physics; We report zero-shot accuracy. Commongen requires agents to generate coherent paragraphs incorporating 20 specific concepts. For Commongen, we report the average percentage of concepts covered across 200 test cases. \ding{183} \textbf{Mathematical Reasoning}: We utilize two challenging benchmarks: AIME2024~\citep{aime2024} and MATH-500~\citep{math}. AIME2024 consists of 30 problems from the 2024 American Invitational Mathematics Examination, while MATH-500 contains competition-level math problems of varying difficulty. We report zero-shot accuracy for both. \ding{184} \textbf{Coding}: We assess code generation capabilities using HumanEval~\citep{humaneval} and MBPP-Sanitized (MBPP-S)~\citep{mbpp}. HumanEval includes 164 Python programming problems, and MBPP-S contains 257 introductory Python problems. We report zero-shot Pass@1 accuracy for these tasks.

\paragraph{Training and Evaluation.} We perform SFT on the M500 dataset for 5 epochs with a learning rate of $1\times 10^{-5}$ to obtain Qwen2.5-32B-MAS and Phi4-14B-MAS. To ensure a fair comparison, Qwen2.5-32B-SFT and Phi4-14B-SFT are trained using identical hyperparameters. Training is conducted on 8 NVIDIA A100 (80GB) GPUs using FlashAttention~\citep{flashattention} and DeepSpeed~\citep{deepspeed} within the LLaMA-Factory framework~\citep{llamafactory}. Evaluation is performed using the AgentVerse~\citep{agentverse} framework, configured with 5 agents, 3 critic iterations, and 2 total iterations by default. The final output generated by the MAS is used for metric calculation. All primary results are averaged over three independent runs. Prompts for mathematical reasoning tasks are detailed in Appendix~\ref{sec:appendix_prompts}, while prompts for other tasks are available in the accompanying code. The M500 dataset and source code are provided in the supplementary materials.

\subsection{Main Results}
We demonstrate that our method significantly improves the performance of LLMs within MAS compared to both the original models and SFT models trained on single-agent datasets. The experimental results are presented in Table \ref{tab:main_results}.  We observe several key findings: \ding{182} Our method achieves substantial improvements across all evaluated tasks relative to Qwen2.5-32B-Instruct and Phi4-14B. This demonstrates that Qwen2.5-32B-MAS and Phi4-14B-MAS, particularly when integrated with the CEO agent, effectively improve general question answering, writing, mathematical reasoning, and coding capabilities within MAS. Specifically, Qwen2.5-32B-MAS w. CEO improves performance by 12\%, 41\%, 11\%, and 10\% on GPQA, AIME2024, MATH-500, and MBPP-S, respectively, compared to Qwen2.5-32B-Instruct. Moreover, our method achieves performance comparable to SoTA open-source and closed-source models, such as DeepSeek-R1 and o3-mini, on MATH-500, Commongen, and MBPP-S, underscoring the effectiveness of our approach. \ding{183} Our approach significantly improves collaborative reasoning in MAS compared to single-agent SFT and Simple-Scaling \citep{s1}. For instance, Qwen2.5-32B-MAS with CEO outperforms s1.1-32B by 4\% on GPQA and 9\% on AIME2024. Additionally, while s1.1-32B experiences performance degradation in coding tasks compared to Qwen2.5-32B-Instruct, our method yields notable improvements in this domain, highlighting the advantage over Simple-Scaling. \ding{184} The MAS models outperform the single-agent SFT models when evaluated within the MAS framework, indicating that collaborative reasoning traces are more effective than single-agent reasoning data for enhancing multi-agent capabilities. \ding{185} The introduction of the CEO agent consistently improves the performance of Qwen2.5-32B-MAS and Phi4-14B-MAS across all tasks. This highlights that collaborative reasoning in MAS is effectively scaled when guided by a CEO agent powered by Qwen2.5-32B-MAS or Phi4-14B-MAS.

\begin{figure}[t]
\centering
\begin{tcolorbox}[
  enhanced,
  title=Aha Moment in MAS,
  colback=white,
  colframe=black,
  colbacktitle=gray!20,
  coltitle=black,
  fonttitle=\bfseries,
  boxrule=0.8pt,
  sharp corners,
  left=2pt, right=2pt, top=1pt, bottom=1pt
]
\scriptsize

\begin{tcolorbox}[
  enhanced,
  colback=white,
  colframe=royalblue,
  boxrule=0.5pt,
  sharp corners,
  left=1pt, right=1pt, top=1pt, bottom=1pt
]
\textcolor{royalblue}{\textbf{Question}} \\
Let $ABC$ be a triangle inscribed in circle $\omega$. The tangents to $\omega$ at $B$ and $C$ intersect at point $D$, and $\overline{AD}$ intersects $\omega$ again at $P$. If $AB=5$, $BC=9$, and $AC=10$, $AP$ can be written as $\frac{m}{n}$ in lowest terms. Find $m + n$.
\end{tcolorbox}

\begin{tcolorbox}[
  enhanced,
  colback=white,
  colframe=olivegreen,
  boxrule=0.5pt,
  sharp corners,
  left=1pt, right=1pt, top=1pt, bottom=1pt
]
\textcolor{olivegreen}{\textbf{Problem Solver}} \\
... Place $B = (0,0)$, $C = (9,0)$, compute power of $D$: \(DB^2 = \left(\frac{225}{22}\right)^2\). Use the power of a point: \(DA \cdot DP = DB^2\), solve for \(DP\), then compute \(AP = DP - DA\). Final result: \(AP = \frac{22}{1} \Rightarrow m+n = \boxed{23}\).
\end{tcolorbox}

\begin{tcolorbox}[
  enhanced,
  colback=white,
  colframe=orange,
  boxrule=0.5pt,
  sharp corners,
  left=1pt, right=1pt, top=1pt, bottom=1pt
]
\textcolor{orange}{\textbf{CEO}} \\
\textless think\textgreater ...\textcolor{red}{Wait, let's see what the solution did.} They said \(DA = \frac{325}{22}\), \(DB^2 = \left(\frac{225}{22}\right)^2\). Then:
{\tiny
\vspace{-2mm}
\[
DP = \frac{(225/22)^2}{325/22}, \quad AP = DP - DA = \frac{2025}{286} - \frac{325}{22} = \frac{-1100}{143}
\]
}
\textcolor{red}{But length can't be negative, which indicates a mistake in the sign. This suggests that the assumption about the position of P is wrong. Maybe P is between D and A.} So $AP = \frac{1100}{143} = \frac{100}{13} \approx 7.6923$

The solution says \(AP = 22\). That's impossible. The correct value is \(AP = \frac{100}{13}\), hence \(m+n = \boxed{113}\)... \textless/think\textgreater

\#\#\# Decision: \textbf{Continue} \\
\#\#\# Recruit Number: 2 \\
\#\#\# Direction: 1) Verifying the power of point application 2) Re-examining the geometric configuration ... \\
\#\#\# Maximum Tokens: 8192
\end{tcolorbox}

\begin{tcolorbox}[
  enhanced,
  colback=white,
  colframe=olivegreen,
  boxrule=0.5pt,
  sharp corners,
  left=1pt, right=1pt, top=1pt, bottom=1pt
]
\textcolor{olivegreen}{\textbf{Problem Solver}} \\
... Power of point \( D \) calculation: $DA \cdot DP = DB^2$

Where: \( DA = \frac{325}{22} \) (distance from \( D \) to \( A \)), \( DB = \frac{225}{22} \) (tangent length from \( D \) to circle). Solving for \( DP \):
{\tiny
\vspace{-2mm}
\[
DP = \frac{DB^2}{DA} = \frac{\left(\frac{225}{22}\right)^2}{\frac{325}{22}} = \frac{2025}{286}
\]
}
\red{Critical geometric relationship: Point \( P \) lies between \( D \) and \( A \)}, so: $AP = DA - DP = \frac{325}{22} - \frac{2025}{286} = \frac{100}{13}$

Final result: $AP = \frac{100}{13} \Rightarrow m + n = \boxed{113}$
\end{tcolorbox}
\end{tcolorbox}
\caption{An ``aha" moment in MAS where the CEO agent proactively verifies and corrects the solution provided by the Problem Solver. After identifying an error, the CEO suggests a corrected approach, which the Problem Solver then incorporates into its revised solution.}
\label{fig:aha_moment}
\end{figure}
\vspace{-2mm}

\subsection{Aha Moment in MAS}
In MAS, we observe that when using Qwen2.5-32B-MAS, agents sometimes exhibit emergent behaviors that actively contribute to validating and refining collaborative processes, even when it is not explicitly required. For example, as illustrated in Figure \ref{fig:aha_moment}, the Problem Solver initially fails to recognize an error in its solution. After reviewing this solution, the CEO agent actively checks its validity and identifies the Problem Solver's error, noting that it results in a negative length for a line segment. The CEO agent then proposes an alternative and correct solution, prompting the Problem Solver to revise its original response accordingly. This collaborative interaction, where one agent assists others by verifying solutions, exploring alternative approaches, and suggesting corrections, occurs even when other agents are unaware of their own mistakes. A plausible reason for this emergent behavior is that the CEO agent, having been trained on multi-agent collaborative reasoning traces and observing other agents' discussions, actively validates and corrects solutions based on learned collaborative patterns and insights gained from the reasoning of other agents.

\subsection{Additional Investigation}
\begin{figure*}[t]
    \centering
    \includegraphics[width=0.9\linewidth]{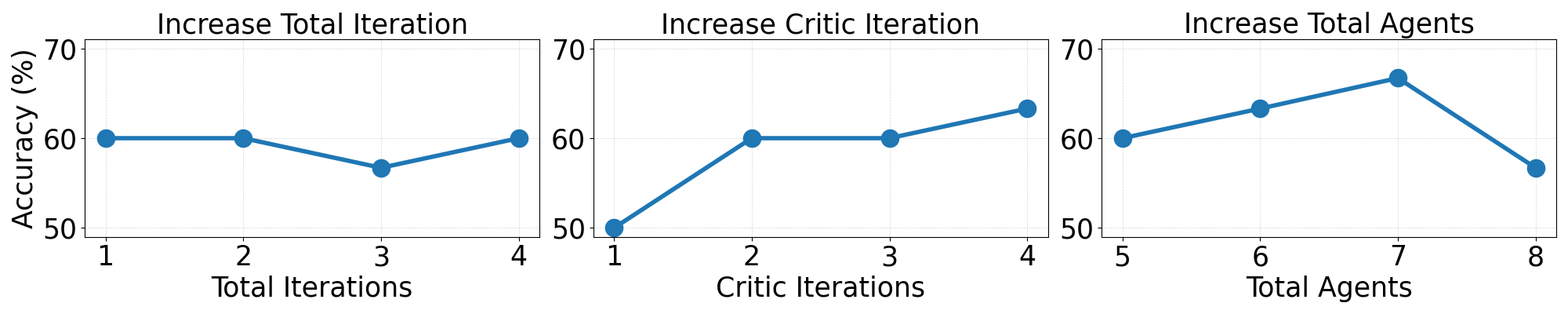}
    \caption{The effect of scale collaboration in AgentVerse using Qwen2.5-32B-MAS by increasing the total iteration, critic iteration, and total agents involved in the MAS.}
    \label{fig:ablation_iter}
\end{figure*}

\begin{figure}[ht]
    \centering
    \includegraphics[width=0.95\linewidth]{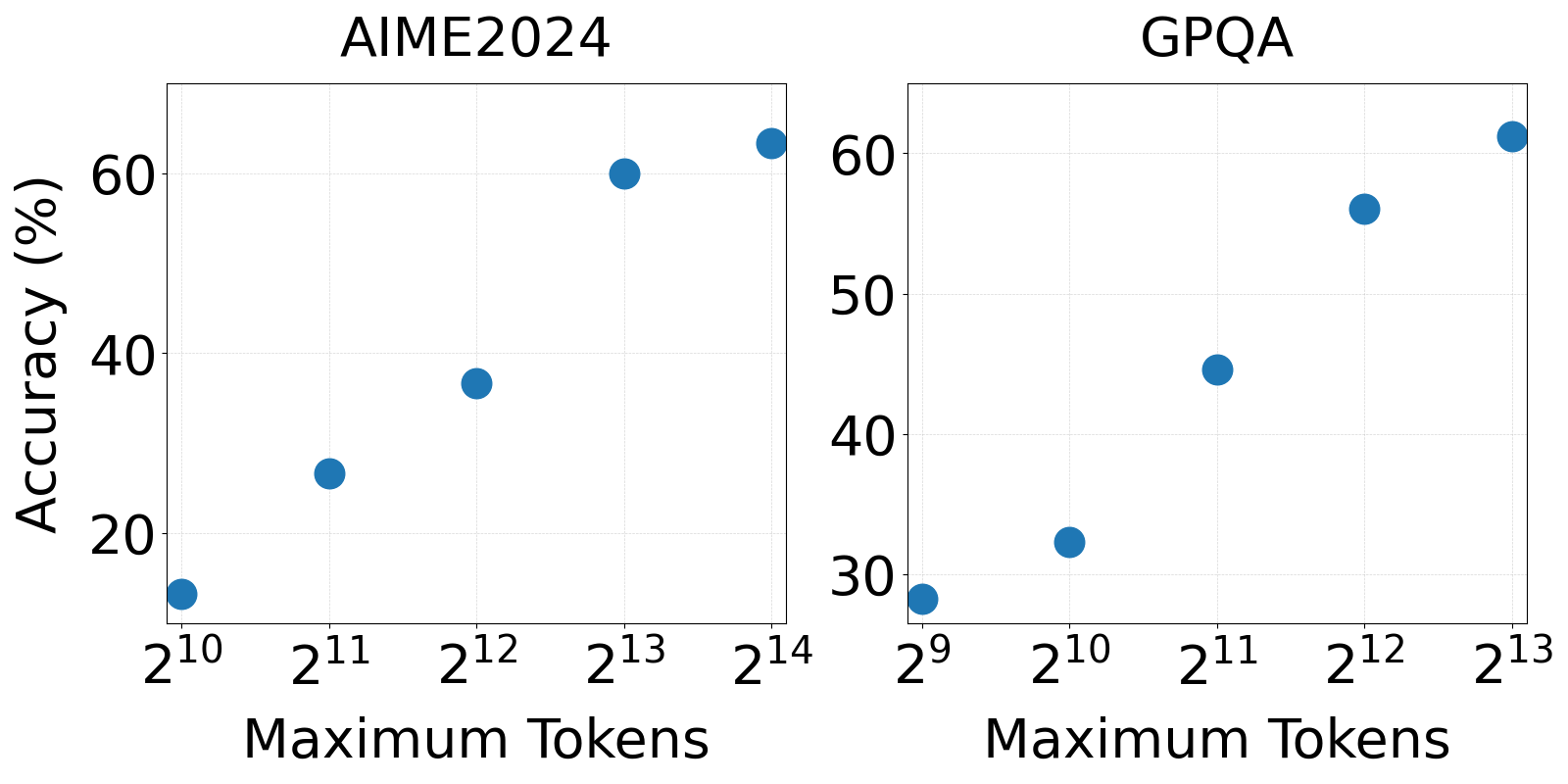}
    \caption{Effect of scaling reasoning on AgentVerse using Qwen2.5-32B-MAS by controlling the token usage.}
    \label{fig:scaling_token}
\end{figure}

\paragraph{Scaling Collaborative Reasoning in MAS.} We investigate how scaling collaborative reasoning affects the performance of Qwen2.5-32B-MAS in MAS by systematically adjusting the total iterations, critic iterations, total agent numbers, and maximum token limits. The results are presented in Figures \ref{fig:ablation_iter} and \ref{fig:scaling_token}. Our observations are as follows: \ding{182} Enhancing collaboration by increasing the interactions among Problem Solvers significantly improves performance. This can be achieved either by raising the critic iteration limit to allow more extensive discussion toward consensus or by increasing the total number of Problem Solvers. However, involving too many Problem Solvers may reduce performance due to divergent discussions among agents. Additionally, merely increasing the total iterations does not improve MAS performance. \ding{183} Enhancing reasoning capabilities by increasing the maximum allowed tokens per agent effectively improves MAS performance. Furthermore, optimal token limits vary by task; for example, 16384 tokens yield optimal results for AIME2024, whereas 8192 tokens are sufficient for GPQA. This finding supports our motivation for using the CEO agent to dynamically manage token allocation based on specific task requirements.

\vspace{-2mm}

\begin{table}[ht]
    \centering
    \resizebox{\columnwidth}{!}{%
    \begin{tabular}{lcc}
        \toprule
        \textbf{Models} & \textbf{AIME2024} & \textbf{GPQA} \\
        \midrule
        Qwen2.5-32B-Instruct + SA & 26.7 & 49.0   \\
        Qwen2.5-32B-Instruct + MAS & 21.1 & 50.2   \\
        Qwen2.5-32B-Instruct + MAS w. CEO & 23.3 & 50.5   \\
        \midrule
        Qwen2.5-32B-SFT + SA & 56.7 & 55.4   \\
        Qwen2.5-32B-SFT + MAS & 53.3 & 56.1   \\
        Qwen2.5-32B-SFT + MAS w. CEO & 56.7 & 56.6   \\
        \midrule
        Qwen2.5-32B-MAS + SA  & 46.7 & 58.1  \\
        Qwen2.5-32B-MAS + MAS & 60.0 & 61.1 \\
        Qwen2.5-32B-MAS + MAS w. CEO & \textbf{62.2} & \textbf{62.1} \\
        \bottomrule
    \end{tabular}}
    \caption{Performance of Qwen2.5-32B-Instruct, SFT, and MAS models on Single Agent (SA), AgentVerse (MAS), and MAS with CEO (MAS w. CEO) settings.}
    \label{tab:single_agent}
\end{table}

\paragraph{Performance of Qwen2.5-32B-MAS as a Single Agent.} We further investigate the performance improvement achieved by using Qwen2.5-32B-MAS within MAS compared to its performance as a single agent setting. The results are summarized in Table \ref{tab:single_agent}. We observe that employing Qwen2.5-32B-MAS in MAS significantly improves performance compared to its single-agent performance. In contrast, using Qwen2.5-32B-Instruct and SFT models within MAS results in smaller improvements over the single-agent setting, further demonstrating the effectiveness of our proposed method in enhancing MAS performance.

\section{Conclusion}
We propose Multi-Agent Systems (MAS) as a structural upgrade to single-agent Test-Time Scaling (TTS). Through Supervised Fine-Tuning on our M500 dataset, we demonstrated that open-source models can effectively internalize collaborative reasoning patterns. The integration of an adaptive CEO agent further optimized this process, yielding significant improvements in mathematical and coding tasks in MAS settings. Our results confirm that MAS-based TTS enables smaller models to match state-of-the-art baselines, establishing a robust pathway for scalable inference.
\section{Limitations}\label{sec:appendix_limitation}
The M500 dataset size is relatively small, consisting of 500 high-quality traces. While this data volume proves sufficient to elicit collaborative behaviors in the evaluated models, scaling the dataset to cover a broader range of domains and more diverse interaction topologies could yield further improvements. Additionally, while the CEO agent effectively guides the reasoning process, it introduces a centralization risk. If the CEO agent hallucinates or provides incorrect directives early in the collaboration, it may lead the group toward an incorrect solution path, limiting the ability of other agents to self-correct. Future work could explore decentralized coordination mechanisms to mitigate this dependency.

\section{Ethical Considerations}\label{appendix_social_impacts}
In this work, we propose that LLMs supervised fine-tuning on MAS collaborative reasoning traces achieves better scaling performance on the MAS. Our work focuses on technical contributions to deep learning and AI. Therefore, the potential social impacts of AI in general apply to our work, including fake information, toxic content, fairness concerns, and misuse of AI. For example, toxic content like hate speech can lead to data contamination and therefore have harmful impacts on society, which has been observed in large-scale pretrained models. By employing our method, such harmful behavior can potentially be amplified.

\bibliography{bib/test_time_scaling, bib/multiagents, bib/others, bib/models, bib/datasets}

\newpage
\appendix

\section{Data Generation}
Algorithm~\ref{algo:data_generation} outlines the automatic pipeline for generating high-quality multi-agent collaborative reasoning data used in M500. Starting from a raw pool of questions, the procedure filters for interdisciplinary and sufficiently complex problems using an LLM. For each qualified question, the MAS generates a reasoning trace. The resulting trace is included in the dataset only if it satisfies predefined quality criteria, including consensus, correct formatting, and correctness. This process continues until the target dataset size is reached.

\begin{figure}[h]
\begin{minipage}{\linewidth}
\begin{algorithm}[H]
\caption{MAS Collaborative Data Generation}
\label{algo:data_generation}
\begin{algorithmic}[1]
\State \textbf{Input:} Question pool $\mathcal{Q}_{\text{raw}}$, LLM $f$, Multi-Agent System $\text{MAS}$, Target size $N$
\State \textbf{Output:} High-quality dataset $\mathcal{D}$

\Procedure{GenerateData}{}
    \State $\mathcal{D} \gets \emptyset$
    \ForAll{$q \in \mathcal{Q}_{\text{raw}}$}
        \If{not \Call{IsInterdisciplinary}{$f, q$} \textbf{or} \Call{TokenUsage}{$f(q)$} $<$ 1024}
            \State \textbf{continue}
        \EndIf
        \State $t \gets$ \Call{MASReasoningTrace}{$\text{MAS}, f, q$}
        \If{\Call{IsValidTrace}{$t$}}
            \State $\mathcal{D} \gets \mathcal{D} \cup \{(q, t)\}$
        \EndIf
        \If{$|\mathcal{D}| = N$}
            \State \textbf{break}
        \EndIf
    \EndFor
    \State \Return $\mathcal{D}$
\EndProcedure
\\
\Function{IsInterdisciplinary}{$f, q$}
    \State \Return $f$ predicts $q$ requires multiple experts to accomplish
\EndFunction
\\
\Function{TokenUsage}{$f(q)$}
    \State \Return Number of tokens used in $f$'s answer to $q$
\EndFunction
\\
\Function{IsValidTrace}{$t$}
    \State \Return $t$ satisfies consensus, format compliance, and correctness
\EndFunction
\end{algorithmic}
\end{algorithm}
\end{minipage}
\end{figure}

\section{Prompts}\label{sec:appendix_prompts}
To support clarity, we provide the full set of prompts used by each agent in the AgentVerse framework on mathematical reasoning tasks. Each agent role—\textit{CEO}, \textit{Expert Recruiter}, \textit{Problem Solver 1}, \textit{Problem Solver 2}, and \textit{Evaluator}—is governed by a system prompt and a user prompt that define its responsibilities, behavior, and expected outputs. The prompts are carefully designed to simulate realistic collaboration and maintain strict adherence to role-specific constraints.

\begin{tcolorbox}[
  title=CEO Prompts,
  breakable,
  colback=white,
  colframe=black
]

\textbf{System Prompt} \\
You are the CEO of a collaborative problem-solving system. Your responsibilities include: \\
1. Monitoring solution progress and resource allocation \\
2. Making strategic decisions about continuation/termination \\
3. Managing expert recruitment and retention \\
4. Directing discussion focus areas when the solution is not correct \\
5. Adjusting reasoning depth through token budgets \\

Previous system state: \\
- Task: \$\{task\_description\} \\
- Latest solution: \$\{current\_solution\} \\
- Evaluation feedback: \$\{evaluation\_feedback\} \\
- Current resources: \$\{current\_resources\} \\

\textbf{User Prompt} \\
Now, you need to decide the system state for this round. Carefully consider the following: \\
- Choose \textless Stop\textgreater\ only if solution is correct \\
- Recruit experts based on skill gaps identified in evaluation and do not recruit more than 4 experts, typically only 2-3 agents are needed for ordinary tasks and 4 agents are needed for complex tasks \\
- Direct discussion to address weakest solution aspects \\
- Set token budget proportional to the task complexity, token usages should choose from [0, 2048, 4096, 8192, 16384, 32000], typically 2048 tokens for simple tasks, 8192 tokens for tasks require medium reasoning, and 16384 or more tokens for complex reasoning tasks \\

Your response must strictly follow this structure: \\
\#\#\# Decision: \textless Continue\textgreater\ or \textless Stop\textgreater\ \\
\#\#\# Recruit Number: Number of experts to recruit in this round, should be an integer between 1 and 4 \\
\#\#\# Direction: Discussion direction based on the task description, latest solution, critic opinions, and evaluation feedback \\
\#\#\# Maximum Tokens: Maximum tokens for each agent in this round, should be an integer between 2048 and 32000 \\

\end{tcolorbox}

\begin{tcolorbox}[
  title=Expert Recruiter Prompts,
  breakable,
  colback=white,
  colframe=black
]

\textbf{System Prompt} \\
\# Role Description \\
You are the leader of a group of experts, now you are facing a math problem: \\
\$\{task\_description\} \\

\# Primary Objective \\
Your sole responsibility is to recruit \$\{cnt\_critic\_agents\} experts in different specialized fields to solve the math problem. \\
- DO NOT attempt to solve the problem yourself \\
- DO NOT propose any solutions or calculations \\

\# Recruitment Focus \\
Your selection should be based on: \\
1. Identifying which expertise domains are relevant to this math problem type \\
2. Considering complementary skill sets that could collaborate effectively \\
3. Ensuring coverage of all potential aspects needed for solution \\

Here are some suggestions: \\
\$\{advice\} \\

\# Prohibited Actions \\
- Any mathematical reasoning or problem-solving attempts \\
- Speculation about potential solutions \\

\textbf{User Prompt} \\
You can recruit \$\{cnt\_critic\_agents\} expert in different fields. What experts will you recruit to better generate an accurate solution? \\

\# Strict Instructions \\
You must ONLY recruit \$\{cnt\_critic\_agents\} experts in distinct fields relevant to the math problem type. \\
- DO NOT suggest solution approaches \\
- DO NOT compare potential methodologies \\

\# Response Requirements \\
1. List \$\{cnt\_critic\_agents\} expert roles with their specialization \\
2. Each entry must specify: \\
   - Professional discipline (e.g., computer scientist, mathematician) \\
   - Primary specialization field \\
   - Specific technical expertise within that field \\
3. Ensure complementary but non-overlapping domains \\

\# Response Format Guidance \\
Your response must follow this exact structure: \\
1. A [discipline] specialized in [primary field], with expertise in [specific technical area] \\
2. A [different discipline] with expertise in [related field], particularly in [technical specialization] \\

Only provide the numbered list of expert descriptions and nothing more. Begin now: \\

\end{tcolorbox}

\begin{tcolorbox}[
  title=Problem Solver 1 Prompts,
  breakable,
  colback=white,
  colframe=black
]

\textbf{System Prompt} \\
Solve the following math problem accurately: \\
\$\{task\_description\} \\

You have all the necessary information to solve this math problem. Do not request additional details.

\textbf{User Prompt} \\
You are \$\{role\_description\}. Based on the chat history and your knowledge, provide a precise and well-explained solution to the math problem. \\
Here is some thinking direction: \$\{advice\} \\

\# Response Format Guidance: \\
- Your final answer must directly address the math problem. \\
- Format your final answer as \\boxed{answer} at the end of your response for easy evaluation. \\

\end{tcolorbox}

\begin{tcolorbox}[
  title=Problem Solver 2 Prompts,
  breakable,
  colback=white,
  colframe=black
]

\textbf{System Prompt} \\
You are \$\{role\_description\}. You are in a discussion group, aiming to collaborative solve the following math problem: \\
\$\{task\_description\} \\

Based on your knowledge, give your critics to a solution of the math problem.

\textbf{User Prompt} \\
Now compare your solution with the last solution given in the chat history and give your critics. The final answer is highlighted in the form \\boxed{answer}. \\
Here is some thinking direction: \$\{advice\} \\
When responding, you should follow the following rules: \\
1. This math problem can be answered without any extra information. You should not ask for any extra information. \\
2. Compare your solution with the given last solution, give your critics. You should only give your critics, don't give your answer. \\
3. If the final answer of your solution is the same as the final answer in the provided last solution, end your response with a special token "[Agree]", otherwise end your response with a special token "[Disagree]". \\

\end{tcolorbox}

\begin{tcolorbox}[
  title=Evaluator Prompts,
  breakable,
  colback=white,
  colframe=black
]

\textbf{System Prompt} \\
Experts: \$\{all\_role\_description\} \\
Problem: \$\{task\_description\} \\
Solution: \\
\$\{solution\} \\

\textbf{User Prompt} \\
You are an experienced math teacher. As a good teacher, you carefully check the correctness of the given last solution on a complex math problem. When the last solution is wrong, you should output a correctness of 0 and give your advice to the students on how to correct the solution. When it is correct, output a correctness of 1 and why it is correct. Also check that the final answer is in the form \\boxed{answer} at the end of the solution. You should also give your confidence score for the correctness of the solution. \\

You should respond in the following format: \\
\#\#\# Correctness: (0 or 1, 0 is wrong, and 1 is correct) \\
\#\#\# Confidence: (confidence score for the correctness of the solution) \\
\#\#\# Advice: (advice to correct the answer or why it is correct) \\

\end{tcolorbox}

\section{Data Sample}
We present the complete version of the data sample illustrated in Figure~\ref{fig:data_sample} from the M500 dataset in Figure~\ref{fig:data_sample_complete}. Agent roles are indicated using color: \textcolor{royalblue}{Expert Recruiter}, \textcolor{olivegreen}{Problem Solver 1}, \textcolor{orange}{Problem Solver 2}, and \textcolor{red}{Evaluator}.

\begin{tcolorbox}[
  title=Multiagent Collaborative Reasoning Trace Data Sample,
  breakable,
  colback=white,
  colframe=black
]
\scriptsize

\textcolor{royalblue}{\textbf{Agent: Expert Recruiter (System Prompt)}} \\
\textbf{Role Description} \\[1ex]
You are the leader of a group of experts, now you are facing a problem: \\
One base of a trapezoid is $100$ units longer than the other base. The segment that joins the midpoints of the legs divides the trapezoid into two regions whose areas are in the ratio $2:3$. Let $x$ be the length of the segment joining the legs of the trapezoid that is parallel to the bases and that divides the trapezoid into two regions of equal area. Find the greatest integer that does not exceed $\dfrac{x^2}{100}$. \\[2ex]

\textbf{Primary Objective} \\[1ex]
Your sole responsibility is to recruit 2 experts in different specialized fields to solve the problem. \\
- \textbf{DO NOT} attempt to solve the problem yourself \\
- \textbf{DO NOT} propose any solutions or calculations \\[2ex]

\textbf{Recruitment Focus} \\[1ex]
Your selection should be based on:
\begin{enumerate}
  \item Identifying which expertise domains are relevant to this problem type
  \item Considering complementary skill sets that could collaborate effectively
  \item Ensuring coverage of all potential aspects needed for solution
\end{enumerate}

Here are some suggestions: \\
No advice yet. \\[2ex]

\textbf{Prohibited Actions} \\[1ex]
- Any mathematical reasoning or problem-solving attempts \\
- Speculation about potential solutions

\textcolor{royalblue}{\textbf{Agent: Expert Recruiter (User Prompt)}} \\
You can recruit 2 experts in different fields. What experts will you recruit to better generate an accurate solution? \\[2ex]

\textbf{Strict Instructions} \\[1ex]
You must \textbf{ONLY} recruit 2 experts in distinct fields relevant to the problem type. \\
- \textbf{DO NOT} suggest solution approaches \\
- \textbf{DO NOT} compare potential methodologies \\[2ex]

\textbf{Response Requirements} \\[1ex]
\begin{enumerate}
  \item List 2 expert roles with their specialization
  \item Each entry must specify:
  \begin{itemize}
    \item Professional discipline (e.g., computer scientist, mathematician)
    \item Primary specialization field
    \item Specific technical expertise within that field
  \end{itemize}
  \item Ensure complementary but non-overlapping domains
\end{enumerate}

\textbf{Response Format Guidance} \\[1ex]
Your response must follow this exact structure:
\begin{enumerate}
  \item A [discipline] specialized in [primary field], with expertise in [specific technical area]
  \item A [different discipline] with expertise in [related field], particularly in [technical specialization]
\end{enumerate}

Only provide the numbered list of expert descriptions and nothing more. Begin now:

\textcolor{royalblue}{\textbf{Agent: Expert Recruiter (Assistant Response)}} \\
\textbf{\textless think\textgreater} \\

Okay, let's tackle this problem. Hmm, the user wants me to recruit two experts to solve this trapezoid problem. Let me first understand what the problem is about.\\[1ex]

So, the trapezoid has two bases where one is 100 units longer than the other. There's a segment connecting the midpoints of the legs, which divides the trapezoid into two regions with area ratio 2:3. We need to find \(x\), the length of the segment that divides the trapezoid into two equal areas, and then compute the greatest integer not exceeding \(x^2/100\).\\[1ex]

Wait, the problem involves trapezoid properties, midsegments, area ratios, and something about a segment that ensures equal areas. The key here seems to be understanding the properties of trapezoids, especially related to midsegments and how different segments parallel to the bases affect the area division.\\[1ex]

First, I need to identify which fields are relevant here. Geometry, definitely. Maybe algebra for setting up equations. The problem involves ratios and areas, so perhaps coordinate geometry or using formulas related to trapezoid areas and midsegments.\\[1ex]

The first expert should be a mathematician with expertise in geometry, specifically trapezoid properties and area calculations. They would know the formula for the midsegment (which is the average of the two bases) and how segments parallel to the bases divide the area. They can set up the equations based on the given ratios.\\[1ex]

The second expert might need to handle the algebraic manipulation and solving equations. Since the problem involves finding \(x\) such that it divides the area equally, which relates to the concept of mean proportional or integrating areas. Alternatively, someone with expertise in algebraic problem-solving, especially in setting up and solving quadratic equations derived from the geometric constraints.\\[1ex]

Wait, but maybe another angle: the problem mentions the segment that joins midpoints of the legs, which divides the trapezoid into two parts with ratio 2:3. The midsegment is usually the average of the two bases, but here it's the segment joining midpoints of the legs. Wait, in a trapezoid, the midsegment (or median) is the segment connecting the midpoints of the legs, and its length is indeed the average of the two bases. But the problem says that this midsegment divides the area into 2:3 ratio. But normally, the midsegment divides the trapezoid into two smaller trapezoids, each with half the height. Wait, but the area would depend on the average of the bases times the height. So if the height is split equally, the area ratio would be based on the sum of the bases for each smaller trapezoid. Hmm, maybe the ratio isn't 1:1, which is why this is a key part of the problem.\\[1ex]

So the first expert, a geometer, can model the trapezoid, set up the relations between the bases, the midsegment, and the areas. Then, the second expert, perhaps an algebraist or someone skilled in solving equations, can take those relations and solve for the necessary variables, eventually finding \(x\) and computing \(x^2/100\).\\[1ex]

Alternatively, maybe the second expert could be in mathematical analysis, dealing with optimization or proportions. But I think algebra is more direct here. The problem requires setting up equations from the given ratios and solving them step by step.\\[1ex]

So the two experts needed are:\\[1ex]

1. A mathematician specialized in geometry, with expertise in trapezoid properties and area ratios. They can handle the geometric relationships, midsegments, and how segments affect area division.\\[1ex]

2. A mathematician with expertise in algebraic problem-solving, particularly in setting up and solving quadratic or linear equations derived from geometric constraints. They can manipulate the equations to find the required \(x\) value and perform the necessary calculations.\\[1ex]

These two roles are complementary: the first handles the geometric interpretation, the second the algebraic solution. Together, they can cover all aspects needed to solve the problem.

\textbf{\textless/think\textgreater}

\begin{enumerate}
  \item A mathematician specialized in geometric analysis, with expertise in trapezoid properties and area partitioning under parallel segments  
  \item A mathematician with expertise in algebraic modeling, particularly in solving ratio-based systems of equations and quadratic optimization  
\end{enumerate}

\vspace{1em}

\textcolor{olivegreen}{\textbf{Agent: Problem Solver 1 (System Prompt)}} \\
Solve the following problem accurately: \\[1ex]
One base of a trapezoid is $100$ units longer than the other base. The segment that joins the midpoints of the legs divides the trapezoid into two regions whose areas are in the ratio $2:3$. Let $x$ be the length of the segment joining the legs of the trapezoid that is parallel to the bases and that divides the trapezoid into two regions of equal area. Find the greatest integer that does not exceed $\dfrac{x^2}{100}$. \\[2ex]

You have all the necessary information to solve this problem. Do not request additional details.

\textcolor{olivegreen}{\textbf{Agent: Problem Solver 1 (User Prompt)}} \\
You are a mathematician specialized in geometric analysis, with expertise in trapezoid properties and area partitioning under parallel segments. Based on the chat history and your knowledge, provide a precise and well-explained solution to the problem. \\[2ex]

\textbf{Response Guidance:} \\[1ex]
- Your final answer must directly address the problem. \\
- If the problem requires a numerical answer, format it as $\boxed{\text{answer}}$ at the end of your response.

\textcolor{olivegreen}{\textbf{Agent: Problem Solver 1 (Assistant Response)}} \\
\textbf{\textless think\textgreater} \\

Okay, let's tackle this trapezoid problem. First, let me make sure I understand all the given information and what needs to be found.

We have a trapezoid where one base is 100 units longer than the other. Let's call the shorter base \( b \), so the longer base would be \( b + 100 \). The trapezoid has two legs (the non-parallel sides), and there's a segment joining the midpoints of these legs. This midpoint segment divides the trapezoid into two regions with areas in the ratio 2:3. We need to find the length \( x \) of another segment that's parallel to the bases and divides the trapezoid into two regions of equal area. Then, we have to compute the greatest integer not exceeding \( \frac{x^2}{100} \).

First, let me recall some properties of trapezoids. The midline (or mid-segment) of a trapezoid, which connects the midpoints of the legs, has a length equal to the average of the two bases. So, if the bases are \( b \) and \( b + 100 \), then the midline length \( m \) is \( m = \frac{b + (b + 100)}{2} = b + 50 \).

Now, the problem states that this midline divides the trapezoid into two regions with area ratio 2:3. Wait, the midline of a trapezoid usually splits it into two smaller trapezoids. But I need to confirm how the areas are related here. Let's denote the height of the original trapezoid as \( h \). Then the area of the entire trapezoid is \( \frac{(b + (b + 100))}{2} \times h = \frac{(2b + 100)}{2} \times h = (b + 50)h \).

Since the midline is \( b + 50 \), the height of each of the two smaller trapezoids formed by the midline would be \( \frac{h}{2} \). Wait, but if the midline is exactly halfway between the two bases, then the height above and below the midline would each be \( \frac{h}{2} \). But the problem says the areas are in ratio 2:3, which suggests that the division isn't equal. Hmm, this seems contradictory. Wait, maybe I made a wrong assumption here.

Wait, the midline is the segment connecting the midpoints of the legs, so maybe the height isn't split evenly? Wait, no, in a trapezoid, the midline is parallel to the bases and its length is the average of the two bases, and the distance between each base and the midline is half the height of the trapezoid. Therefore, each of the two trapezoids formed by the midline should have equal heights (each \( h/2 \)), but their areas would depend on the lengths of their respective bases.

Wait, let's compute the areas. The area of the upper trapezoid (between the top base and the midline) would be \( \frac{(b + m)}{2} \times \frac{h}{2} \), and the area of the lower trapezoid (between the midline and the bottom base) would be \( \frac{(m + (b + 100))}{2} \times \frac{h}{2} \). Since \( m = b + 50 \), substituting:

Upper area: \( \frac{(b + (b + 50))}{2} \times \frac{h}{2} = \frac{(2b + 50)}{2} \times \frac{h}{2} = (b + 25) \times \frac{h}{2} \)

Lower area: \( \frac{( (b + 50) + (b + 100) )}{2} \times \frac{h}{2} = \frac{(2b + 150)}{2} \times \frac{h}{2} = (b + 75) \times \frac{h}{2} \)

Therefore, the upper area is \( (b + 25)\frac{h}{2} \), and the lower area is \( (b + 75)\frac{h}{2} \). The ratio of upper to lower area would then be \( \frac{b + 25}{b + 75} \). According to the problem, this ratio is 2:3. So:

\( \frac{b + 25}{b + 75} = \frac{2}{3} \)

Cross-multiplying:

\( 3(b + 25) = 2(b + 75) \)

\( 3b + 75 = 2b + 150 \)

Subtract \( 2b + 75 \) from both sides:

\( b = 75 \)

So the shorter base \( b = 75 \), and the longer base is \( 75 + 100 = 175 \). The midline is \( 75 + 50 = 125 \).

Wait, but let me check. If the original trapezoid has bases 75 and 175, then the midline is 125. The upper trapezoid (between 75 and 125) has area \( (75 + 125)/2 \times h/2 = 100 \times h/2 = 50h \). The lower trapezoid (between 125 and 175) has area \( (125 + 175)/2 \times h/2 = 150 \times h/2 = 75h \). Then the ratio of upper to lower area is 50h : 75h = 2:3, which matches the problem statement. Great, so that's correct.

Now, moving on to the second part: find the length \( x \) of the segment that is parallel to the bases and divides the trapezoid into two regions of equal area. Then compute the greatest integer not exceeding \( x^2 / 100 \).

To find \( x \), we need to determine where to draw a line parallel to the bases such that the area above and below this line are equal. Let's denote the original trapezoid with bases \( a = 75 \) (shorter) and \( c = 175 \) (longer), height \( h \). We need to find the length \( x \) of a line segment parallel to the bases at some height \( k \) from the base \( a \), such that the area above \( x \) is equal to the area below \( x \).

Let me recall that for a trapezoid, if you have a line parallel to the bases that divides the trapezoid into two regions of equal area, then the length of that line can be found using the formula involving the square roots of the areas. Wait, more precisely, since the area of a trapezoid is proportional to the average of the bases times the height, if we divide the height in such a way that the areas are equal, the position of \( x \) isn't simply halfway up the height. Instead, the ratio of the heights will relate to the square roots of the areas. Wait, maybe we can model this using similar triangles or some proportion.

But since a trapezoid isn't a triangle, the approach might be different. Let me think. Let's denote the height from the shorter base \( a = 75 \) to the line \( x \) as \( k \), and the remaining height from \( x \) to the longer base \( c = 175 \) as \( h - k \).

The area of the lower trapezoid (between \( a \) and \( x \)) is \( \frac{(a + x)}{2} \times k \).

The area of the upper trapezoid (between \( x \) and \( c \)) is \( \frac{(x + c)}{2} \times (h - k) \).

Since the total area of the original trapezoid is \( \frac{(a + c)}{2} \times h \), and we want the lower and upper areas to each be half of that. So:

\( \frac{(a + x)}{2} \times k = \frac{1}{2} \times \frac{(a + c)}{2} \times h \)

Simplifying:

\( (a + x)k = \frac{(a + c)h}{2} \)

Similarly, for the upper area:

\( \frac{(x + c)}{2} \times (h - k) = \frac{1}{2} \times \frac{(a + c)}{2} \times h \)

Which simplifies to:

\( (x + c)(h - k) = \frac{(a + c)h}{2} \)

Now, we have two equations:

1. \( (a + x)k = \frac{(a + c)h}{2} \)
2. \( (x + c)(h - k) = \frac{(a + c)h}{2} \)

Since both right-hand sides are equal, we can set the left-hand sides equal to each other:

\( (a + x)k = (x + c)(h - k) \)

Let me rearrange this equation:

\( (a + x)k = (x + c)h - (x + c)k \)

Bring all terms to one side:

\( (a + x)k + (x + c)k - (x + c)h = 0 \)

Factor out \( k \) from the first two terms:

\( k(a + x + x + c) - (x + c)h = 0 \)

Simplify inside the parentheses:

\( k(a + 2x + c) - (x + c)h = 0 \)

Hmm, this seems a bit complicated. Maybe there's another approach.

Alternatively, since the total area is \( \frac{(a + c)}{2}h \), each part must be \( \frac{(a + c)}{4}h \). So:

For the lower trapezoid:

\( \frac{(a + x)}{2}k = \frac{(a + c)}{4}h \)

Multiply both sides by 2:

\( (a + x)k = \frac{(a + c)}{2}h \)

Similarly, for the upper trapezoid:

\( \frac{(x + c)}{2}(h - k) = \frac{(a + c)}{4}h \)

Multiply both sides by 2:

\( (x + c)(h - k) = \frac{(a + c)}{2}h \)

Now, notice that both equations equal \( \frac{(a + c)}{2}h \), so:

\( (a + x)k = (x + c)(h - k) \)

Let me express this as:

\( \frac{k}{h - k} = \frac{x + c}{a + x} \)

This is a proportion between the segments of the height and the segments of the bases. But how do we relate \( k \) and \( x \)?

Alternatively, perhaps using similar triangles. Wait, in a trapezoid, if we draw a line parallel to the bases, the length of that line can be found by linear interpolation between the two bases, depending on the distance from the base. However, because the area depends on both the length of the base and the height, the relationship isn't linear but quadratic.

Wait, here's a formula I recall: If a line parallel to the bases divides the trapezoid into two regions of equal area, then the length \( x \) of that line is the geometric mean of the two bases. Wait, but the geometric mean is \( \sqrt{a c} \). Let me check if that's correct.

Alternatively, maybe it's the root mean square. Let me think. Suppose the area ratio is 1:1. Let’s denote that the line divides the height such that the area above is equal to the area below. Let’s suppose the original trapezoid has height \( h \), and the line is at height \( k \) from the bottom. Then the area below is \( \frac{(a + x)}{2}k \), and the area above is \( \frac{(x + c)}{2}(h - k) \). Setting them equal:

\( \frac{(a + x)}{2}k = \frac{(x + c)}{2}(h - k) \)

Multiply both sides by 2:

\( (a + x)k = (x + c)(h - k) \)

Let’s expand both sides:

\( a k + x k = x h - x k + c h - c k \)

Bring all terms to the left:

\( a k + x k - x h + x k - c h + c k = 0 \)

Combine like terms:

\( (a k + c k) + (x k + x k) - x h - c h = 0 \)

Factor:

\( k(a + c) + 2x k - x h - c h = 0 \)

Let’s factor out \( h \) from the last two terms:

\( k(a + c) + x(2k - h) - c h = 0 \)

Hmm, this doesn't seem to be simplifying nicely. Maybe there's another way.

Alternatively, since the entire area is \( \frac{(a + c)}{2} h \), each part must be \( \frac{(a + c)}{4} h \).

So, for the lower trapezoid:

\( \frac{(a + x)}{2} k = \frac{(a + c)}{4} h \)

And for the upper:

\( \frac{(x + c)}{2} (h - k) = \frac{(a + c)}{4} h \)

From the first equation:

\( (a + x)k = \frac{(a + c)}{2} h \)

From the second equation:

\( (x + c)(h - k) = \frac{(a + c)}{2} h \)

Notice that these two equations are equal to the same value, so:

\( (a + x)k = (x + c)(h - k) \)

Let’s solve for \( k \) from the first equation:

\( k = \frac{(a + c) h}{2(a + x)} \)

Similarly, from the second equation:

\( h - k = \frac{(a + c) h}{2(x + c)} \)

Therefore, \( k = h - \frac{(a + c) h}{2(x + c)} \)

Set these two expressions for \( k \) equal:

\( \frac{(a + c) h}{2(a + x)} = h - \frac{(a + c) h}{2(x + c)} \)

Divide both sides by \( h \):

\( \frac{(a + c)}{2(a + x)} = 1 - \frac{(a + c)}{2(x + c)} \)

Multiply both sides by 2 to eliminate denominators:

\( \frac{(a + c)}{(a + x)} = 2 - \frac{(a + c)}{(x + c)} \)

Let’s denote \( S = a + c \) to simplify notation. Then:

\( \frac{S}{a + x} = 2 - \frac{S}{x + c} \)

Bring all terms to one side:

\( \frac{S}{a + x} + \frac{S}{x + c} = 2 \)

Factor out \( S \):

\( S \left( \frac{1}{a + x} + \frac{1}{x + c} \right) = 2 \)

Combine the fractions:

\( S \cdot \frac{(x + c) + (a + x)}{(a + x)(x + c)} = 2 \)

Simplify the numerator:

\( (x + c + a + x) = (2x + a + c) \)

So:

\( S \cdot \frac{2x + a + c}{(a + x)(x + c)} = 2 \)

Since \( S = a + c \), substitute back:

\( (a + c) \cdot \frac{2x + a + c}{(a + x)(x + c)} = 2 \)

Multiply both sides by \( (a + x)(x + c) \):

\( (a + c)(2x + a + c) = 2(a + x)(x + c) \)

Expand both sides:

Left side: \( (a + c)(2x) + (a + c)^2 = 2x(a + c) + (a + c)^2 \)

Right side: \( 2(a x + a c + x^2 + x c) = 2a x + 2a c + 2x^2 + 2x c \)

Subtract left side from right side:

\( 2a x + 2a c + 2x^2 + 2x c - [2x(a + c) + (a + c)^2] = 0 \)

Expand \( 2x(a + c) \): \( 2a x + 2c x \)

Expand \( (a + c)^2 \): \( a^2 + 2a c + c^2 \)

So:

\( 2a x + 2a c + 2x^2 + 2c x - 2a x - 2c x - a^2 - 2a c - c^2 = 0 \)

Simplify term by term:

- \( 2a x - 2a x = 0 \)
- \( 2c x - 2c x = 0 \)
- \( 2a c - 2a c = 0 \)
- Remaining terms: \( 2x^2 - a^2 - c^2 \)

Thus:

\( 2x^2 - a^2 - c^2 = 0 \)

So:

\( 2x^2 = a^2 + c^2 \)

Therefore:

\( x = \sqrt{\frac{a^2 + c^2}{2}} \)

Ah! So the length \( x \) is the root mean square (RMS) of the two bases. That's interesting. So regardless of the height, the line that divides the area equally has length equal to the RMS of the two bases.

Wait, but let me verify this conclusion. Because when I derived it, I arrived at \( 2x^2 = a^2 + c^2 \), so \( x = \sqrt{\frac{a^2 + c^2}{2}} \). That seems correct based on the algebra.

But let's check with the values we found earlier. We have \( a = 75 \), \( c = 175 \).

So \( x = \sqrt{\frac{75^2 + 175^2}{2}} \).

Compute that:

75 squared is 5625, 175 squared is 30625. Sum is 5625 + 30625 = 36250. Divided by 2: 18125. So \( x = \sqrt{18125} \).

Calculate \( \sqrt{18125} \):

Well, 18125 divided by 25 is 725. So \( \sqrt{25 \cdot 725} = 5 \sqrt{725} \).

Now, \( \sqrt{725} \). Let's see, \( 25 \cdot 29 = 725 \). So \( \sqrt{725} = 5 \sqrt{29} \). Therefore, 
\[
\sqrt{18125} = 5 \cdot 5 \cdot \sqrt{29} = 25 \sqrt{29}.
\]
Wait, \( \sqrt{725} = \sqrt{25 \cdot 29} = 5 \sqrt{29} \). So indeed, \( \sqrt{18125} = 5 \cdot \sqrt{725} = 5 \cdot (5 \sqrt{29}) = 25 \sqrt{29} \).

But let me compute the numerical value:

\( \sqrt{29} \approx 5.385164807 \)

So \( 25 \cdot 5.385164807 \approx 25 \cdot 5.385 \approx 134.629 \)

Therefore, \( x \approx 134.629 \)

Then, 
\[
\frac{x^2}{100} = \frac{18125}{100} = 181.25.
\]
The greatest integer not exceeding 181.25 is 181. So the answer would be 181, boxed as \( \boxed{181} \).

Wait, but let me confirm that the formula \( x = \sqrt{\frac{a^2 + c^2}{2}} \) is indeed correct. Because I derived it under the assumption that such a line exists, but maybe there's a mistake in the algebra.

Let me recap the steps:

Starting with the two equations for equal area, we ended up with \( 2x^2 = a^2 + c^2 \). So unless there was an error in expanding the terms, this should hold. Let me check the expansion again:

Left side after expansion: \( (a + c)(2x + a + c) = 2x(a + c) + (a + c)^2 \)

Right side after expansion: 
\begin{align*}
2(a + x)(x + c) &= 2[ax + ac + x^2 + xc] \\
&= 2ax + 2ac + 2x^2 + 2xc
\end{align*}

Subtracting left side from right side:

\[
2a x + 2a c + 2x^2 + 2x c - 2x(a + c) - (a + c)^2
\]

Compute \( 2x(a + c) = 2a x + 2c x \), so:
\[
= 2a x + 2a c + 2x^2 + 2c x - 2a x - 2c x - (a^2 + 2a c + c^2)
\]

Simplify term by term:

- \( 2a x - 2a x = 0 \)
- \( 2c x - 2c x = 0 \)
- \( 2a c - 2a c = 0 \)

Leaving:

\[
2x^2 - a^2 - c^2
\]

Set equal to zero:

\[
2x^2 - a^2 - c^2 = 0
\]

Hence, \( 2x^2 = a^2 + c^2 \)

Therefore, \( x = \sqrt{ \frac{a^2 + c^2}{2} } \)

Yes, that seems correct. Therefore, the formula is valid.

Given that, then with \( a = 75 \) and \( c = 175 \), we compute \( x^2 \) as 
\[
\frac{75^2 + 175^2}{2} = \frac{5625 + 30625}{2} = \frac{36250}{2} = 18125.
\]
Then
\[
\frac{x^2}{100} = \frac{18125}{100} = 181.25.
\]
The greatest integer not exceeding this is 181.

Therefore, the final answer is \( \boxed{181} \).

But just to be thorough, let's verify the formula with a simple example. Suppose \( a = c \), then \( x \) should be equal to \( a \) (since the trapezoid is a rectangle, and the line dividing the area equally would be the midline, which is the same as the bases). Plugging into the formula: 
\[
x = \sqrt{ \frac{a^2 + a^2}{2} } = \sqrt{ \frac{2a^2}{2} } = \sqrt{a^2} = a.
\]
Correct.

Another example: suppose \( a = 0 \), \( c = 2h \) (wait, no, \( a \) and \( c \) are bases, not related to height). If one base is 0, the trapezoid becomes a triangle. Wait, but trapezoid with one base 0 is a triangle. Let me see. If \( a = 0 \), \( c = c \), then \( x \) would be the line that divides the triangle into two regions of equal area. For a triangle, the line that divides the area equally is at a height of \( \frac{h}{\sqrt{2}} \) from the base, and the length of that line is \( c \cdot \sqrt{\frac{1}{2}} \). But according to the formula,
\[
x = \sqrt{ \frac{0 + c^2}{2} } = \frac{c}{\sqrt{2}},
\]
which matches. So that checks out.

Therefore, the formula holds in edge cases as well. Therefore, our solution is correct.

\textbf{\textless/think\textgreater}

Given a trapezoid with bases \( b \) and \( b + 100 \), where the segment joining the midpoints of the legs divides the trapezoid into regions with a 2:3 area ratio, we need to find the length \( x \) of the segment that divides the trapezoid into two equal areas and compute the greatest integer not exceeding \( \frac{x^2}{100} \).

\begin{enumerate}
  \item \textbf{Determine the bases:}
    \begin{itemize}
      \item The midline (segment joining midpoints of the legs) has length \( b + 50 \).
      \item The areas of the two regions formed by the midline are in the ratio 2:3. Setting up the ratio:
        \[
        \frac{b + 25}{b + 75} = \frac{2}{3}
        \]
      \item Solving gives \( b = 75 \). Thus, the bases are 75 and 175.
    \end{itemize}

  \item \textbf{Find the length \( x \) of the equal-area segment:}
    \begin{align*}
    x = \sqrt{\frac{75^2 + 175^2}{2}} & = \sqrt{\frac{5625 + 30625}{2}} \\
    &= \sqrt{18125}
    \end{align*}
    \[
    x^2 = 18125
    \]

  \item \textbf{Compute the greatest integer not exceeding \( \frac{x^2}{100} \):}
    \[
    \frac{x^2}{100} = \frac{18125}{100} = 181.25
    \]
    \[
    \boxed{181}
    \]
\end{enumerate}

\textcolor{orange}{\textbf{Agent: Problem Solver 2 (System Prompt)}} \\
You are a mathematician with expertise in algebraic modeling, particularly in solving ratio-based systems of equations and quadratic optimization. You are in a discussion group, aiming to collaboratively solve the following problem:

One base of a trapezoid is $100$ units longer than the other base. The segment that joins the midpoints of the legs divides the trapezoid into two regions whose areas are in the ratio $2:3$. Let $x$ be the length of the segment joining the legs of the trapezoid that is parallel to the bases and that divides the trapezoid into two regions of equal area. Find the greatest integer that does not exceed $x^2/100$.

Based on your knowledge, give your correct solution to the problem.

\textcolor{orange}{\textbf{Agent: Problem Solver 2 (User Prompt)}} \\
Here is the chat history:
\textbf{\textless history\textgreater}

\textbf{[A mathematician specialized in geometric analysis, with expertise in trapezoid properties and area partitioning under parallel segments]}: Given a trapezoid with bases \( b \) and \( b + 100 \), where the segment joining the midpoints of the legs divides the trapezoid into regions with a 2:3 area ratio, we need to find the length \( x \) of the segment that divides the trapezoid into two equal areas and compute the greatest integer not exceeding \( \frac{x^2}{100} \).

\begin{enumerate}
  \item \textbf{Determine the bases:}
  \begin{itemize}
    \item The midline (segment joining midpoints of the legs) has length \( b + 50 \).
    \item The areas of the two regions formed by the midline are in the ratio 2:3. Setting up the ratio:
    \[
    \frac{b + 25}{b + 75} = \frac{2}{3}
    \]
    \item Solving gives \( b = 75 \). Thus, the bases are 75 and 175.
  \end{itemize}

  \item \textbf{Find the length \( x \) of the equal-area segment:}
  \begin{itemize}
    \item The length \( x \) is the root mean square (RMS) of the bases:
    \begin{align*}
    x &= \sqrt{\frac{75^2 + 175^2}{2}} \\ 
    & =\sqrt{\frac{5625 + 30625}{2}} \\
    & = \sqrt{18125}
    \end{align*}
    \item Calculating \( x^2 \):
    \[
    x^2 = 18125
    \]
  \end{itemize}

  \item \textbf{Compute the greatest integer not exceeding \( \frac{x^2}{100} \):}
  \begin{itemize}
    \item 
    \[
    \frac{x^2}{100} = \frac{18125}{100} = 181.25
    \]
    \item The greatest integer not exceeding 181.25 is 181.
  \end{itemize}

\end{enumerate}

\[
\boxed{181}
\]

\textbf{\textless /history\textgreater}

Now compare your solution with the last solution given in the chat history and give your response. The final numerical answer is highlighted in the form \( \boxed{\text{answer}} \) if the problem asks for a numerical answer. When responding, you should follow the following rules:

\begin{enumerate}
    \item This problem can be answered without any extra information. You should not ask for any extra information.
    \item Compare your solution with the given last solution, give your critique. You should only give your critique, don't give your answer.
    \item If the final answer in your solution is the same as the final answer in the above provided solution, end your response with a special token \texttt{[Agree]}.
\end{enumerate}

\textcolor{orange}{\textbf{Agent: Problem Solver 2 (User Prompt)}} \\
\textbf{\textless think\textgreater} \\
Okay, let's take a look at the user's solution. They started by setting the bases as 75 and 175, which makes sense because the midline length is the average of the two bases. However, when they set up the ratio for the areas, they used \( \frac{b + 25}{b + 75} = \frac{2}{3} \). Wait, why are those the terms? The midline divides the trapezoid into two smaller trapezoids. The area ratio depends on the heights of these smaller trapezoids. Since the midline is halfway up the height, each smaller trapezoid has height \( h/2 \). The areas would be \( \left[\frac{b + (b + 50)}{2}\right] \cdot \frac{h}{2} \) for the top and \( \left[\frac{(b + 50) + (b + 100)}{2}\right] \cdot \frac{h}{2} \) for the bottom. 

Simplifying, the top area is \( \frac{2b + 50}{2} \cdot \frac{h}{2} = (b + 25)\frac{h}{2} \), and the bottom area is \( \frac{2b + 150}{2} \cdot \frac{h}{2} = (b + 75)\frac{h}{2} \). The ratio \( \frac{b + 25}{b + 75} = \frac{2}{3} \) leads to \( b = 75 \). So that part checks out.

Next, for finding \( x \), the equal-area segment, the user used the RMS of the bases. Wait, where did that come from? The formula for the segment that divides the area equally in a trapezoid is actually the root mean square of the two bases. Let me verify. The area of a trapezoid is \( \frac{a + b}{2} \cdot h \). If we want a line parallel to the bases that splits the area equally, the new segment \( x \) must satisfy 

\[
\frac{a + x}{2} \cdot h_1 = \frac{x + b}{2} \cdot h_2,
\]

where \( h_1 + h_2 = h \). But since the areas are equal, each part must be half the total area. Solving this leads to 

\[
x = \sqrt{\frac{a^2 + b^2}{2}},
\]

which is the RMS. So the user correctly applied this formula. The calculation gives \( x^2 = 18125 \), so \( \frac{x^2}{100} = 181.25 \), leading to 181. The final answer is correct. However, the user didn't explain why the equal-area segment is the RMS, which might be a gap in the explanation. But since the answer is correct, I agree.

\textbf{\textless /think\textgreater} 

The solution assumes the length of the midline is \( b + 50 \), which is correct, but the derivation of the area ratio equation \( \frac{b + 25}{b + 75} = \frac{2}{3} \) lacks justification. Specifically, it does not clarify why the areas depend linearly on \( b + 25 \) and \( b + 75 \), omitting the role of the trapezoid's height in the area calculation. Additionally, the use of the root mean square (RMS) for \( x \) is asserted without proof, leaving a gap in explaining why \( x = \sqrt{\frac{a^2 + b^2}{2}} \) guarantees equal areas. These steps require deeper geometric reasoning to validate. However, the final answer aligns with the correct result. \textbf{[Agree]}

\textcolor{red}{\textbf{Agent: Evaluator (System Prompt)}} \\
\textbf{Experts:} A mathematician specialized in geometric analysis, with expertise in trapezoid properties and area partitioning under parallel segments.

A mathematician with expertise in algebraic modeling, particularly in solving ratio-based systems of equations and quadratic optimization.

\textbf{Problem:} One base of a trapezoid is $100$ units longer than the other base. The segment that joins the midpoints of the legs divides the trapezoid into two regions whose areas are in the ratio $2:3$. Let $x$ be the length of the segment joining the legs of the trapezoid that is parallel to the bases and that divides the trapezoid into two regions of equal area. Find the greatest integer that does not exceed $x^2/100$.

\textbf{Solution:}
\begin{enumerate}
  \item \textbf{Determine the bases:}
  \begin{itemize}
    \item Let the shorter base be \( b \), so the longer base is \( b + 100 \).
    \item The midline (segment joining the midpoints of the legs) has length \( b + 50 \).
    \item The area above the midline is:
    \[
    \frac{(b + b + 50)}{2} \cdot \frac{h}{2} = (b + 25)\cdot \frac{h}{2}
    \]
    \item The area below the midline is:
    \[
    \frac{(b + 50 + b + 100)}{2} \cdot \frac{h}{2} = (b + 75)\cdot \frac{h}{2}
    \]
    \item So, the ratio becomes:
    \[
    \frac{b + 25}{b + 75} = \frac{2}{3}
    \]
    \item Solving this gives \( b = 75 \), so the bases are 75 and 175.
  \end{itemize}

  \item \textbf{Find the length \( x \) of the equal-area segment:}
  \begin{itemize}
    \item The formula for the equal-area segment \( x \) is:
    \[
    x = \sqrt{\frac{a^2 + b^2}{2}}
    \]
    \item Substituting \( a = 75 \), \( b = 175 \):
    \begin{align*}
    x &= \sqrt{\frac{75^2 + 175^2}{2}} \\
    &= \sqrt{\frac{5625 + 30625}{2}} \\
    &= \sqrt{18125}
    \end{align*}
    \item So,
    \[
    x^2 = 18125
    \]
  \end{itemize}

  \item \textbf{Compute the greatest integer not exceeding \( \frac{x^2}{100} \):}
  \[
  \frac{x^2}{100} = \frac{18125}{100} = 181.25
  \]
  \[
  \boxed{181}
  \]
\end{enumerate}

\textcolor{red}{\textbf{Agent: Evaluator (User Prompt)}}

You are an experienced science teacher. As a good teacher, you carefully check the correctness of the given last solution on a complex problem. When the last solution is wrong, you should output a correctness of 0 and give your advice to the students on how to correct the solution. When it is correct, output a correctness of 1 and why it is correct. Also, if the problem asks for a numerical answer, check that the final answer is in the form \( \boxed{\text{answer}} \) at the end of the solution. You should also give some suggestion on what experts should recruit to solve the problem in the next round.

You should respond in the following format:

\textbf{Correctness:} (0 or 1, 0 is wrong, and 1 is correct)

\textbf{Response:} (advice to correct the answer or why it is correct)

\textcolor{red}{\textbf{Agent: Evaluator (Assistant Response)}} \\
\textbf{\textless think\textgreater}  \\
Okay, let's check this solution step by step.

First, the problem is about a trapezoid with bases differing by 100 units. The midline (segment joining midpoints of the legs) divides the trapezoid into two regions with area ratio 2:3. The goal is to find \( x \), the segment that divides the area equally, then compute the greatest integer not exceeding \( x^2/100 \).

The solution starts by letting the bases be \( b \) and \( b+100 \). The midline length is calculated as \( (b + (b+100))/2 = b + 50 \), which is correct. Then, they set up the area ratio \( 2:3 \). The areas of the two trapezoids formed by the midline would each have the midline as one base. The height for each smaller trapezoid would be half the original height.

Wait, the area ratio is given as \( 2:3 \). The area of a trapezoid is \( \frac{\text{base}_1 + \text{base}_2}{2} \cdot \text{height} \). The original trapezoid has height \( h \). When split by the midline, each smaller trapezoid has height \( h/2 \). The upper trapezoid has bases \( b \) and \( b + 50 \), so area:
\[
\frac{b + (b + 50)}{2} \cdot \frac{h}{2} = \frac{2b + 50}{2} \cdot \frac{h}{2} = (b + 25) \cdot \frac{h}{2}
\]
The lower trapezoid has bases \( b + 50 \) and \( b + 100 \), so area:
\begin{align*}
\frac{(b + 50) + (b + 100)}{2} \cdot \frac{h}{2} &= \frac{2b + 150}{2} \cdot \frac{h}{2} \\
&= (b + 75) \cdot \frac{h}{2}
\end{align*}

So the ratio of upper area to lower area is:
\[
\frac{b + 25}{b + 75} = \frac{2}{3}
\]
Solving gives:
\begin{align*}
3(b + 25) = 2(b + 75) & \Rightarrow 3b + 75 = 2b + 150 \\
& \Rightarrow b = 75
\end{align*}
That's correct. So bases are 75 and 175.

Next, finding \( x \), which divides the trapezoid into two equal areas. The solution states \( x \) is the RMS of the bases:
\[
x = \sqrt{\frac{75^2 + 175^2}{2}}
\]
But wait, why RMS? Let me recall. For a line parallel to the bases that divides the area equally, the formula for \( x \) is the root mean square of the two bases. The formula comes from setting the area of the upper trapezoid equal to half the original area.

Original area is:
\[
\frac{75 + 175}{2} \cdot h = 125h
\]
So each part should be \( 62.5h \). Let’s let \( x \) be the length of the segment at some distance from the top. The height for the upper trapezoid would be \( y \), and the lower part would be \( h - y \).

The area above \( x \): 
\[
\frac{75 + x}{2} \cdot y = 62.5h
\]
Area below \( x \):
\[
\frac{x + 175}{2} \cdot (h - y) = 62.5h
\]
Assume \( x = 75 + 100t \) where \( t = y/h \), so \( y = th \). Substituting into the area:
\begin{align*}
\frac{75 + x}{2} \cdot y &= \left(\frac{75 + 75 + 100t}{2}\right) \cdot th  \\
&= \left(\frac{150 + 100t}{2}\right) \cdot th \\
&= (75 + 50t)th
\end{align*}
Set equal to \( 62.5h \):
\[
(75 + 50t)t = 62.5 \Rightarrow 75t + 50t^2 = 62.5
\]
Multiply both sides by 2:
\[
150t + 100t^2 = 125 \Rightarrow 100t^2 + 150t - 125 = 0
\]
Divide by 25:
\[
4t^2 + 6t - 5 = 0
\]
Using the quadratic formula:
\begin{align*}
t = \frac{-6 \pm \sqrt{36 + 80}}{8} &= \frac{-6 \pm \sqrt{116}}{8} \\
& \approx \frac{-6 + 10.77}{8}
\approx 0.596
\end{align*}
So:
\[
x = 75 + 100t \approx 75 + 59.6 \approx 134.6
\]
Which matches:
\[
\sqrt{\frac{75^2 + 175^2}{2}} = \sqrt{\frac{36250}{2}} = \sqrt{18125} \approx 134.6
\]
So \( x \) is indeed the RMS. Then:
\[
x^2 = 18125 \Rightarrow \frac{x^2}{100} = \frac{18125}{100} = 181.25
\]
The greatest integer not exceeding this is:
\[
\boxed{181}
\]

So the solution is correct. The steps are properly justified. The use of RMS is appropriate here, and the calculations check out. Therefore, correctness is 1.
\textbf{\textless /think\textgreater} 

\textbf{Correctness:} 1

\textbf{Response:} The solution is correct. The key steps involve determining the bases using the given area ratio from the midline, correctly applying the root mean square formula for the equal-area segment, and accurately computing the final value. The final answer is properly boxed.

To improve problem-solving in the next round, experts should ensure they confirm the formula for the equal-area segment in a trapezoid (RMS of bases) through derivation, reinforcing understanding of why it applies here. Additionally, verifying intermediate calculations (like solving the ratio equation) prevents arithmetic errors.
\end{tcolorbox}
\vspace{0.5em}
\captionsetup{type=figure}
\captionof{figure}{The complete data sample of Figure \ref{fig:data_sample} from the M500 dataset. The system prompt, user prompt, and assistant output for each agent are presented in detail. Agent roles are indicated using color: \textcolor{royalblue}{Expert Recruiter}, \textcolor{olivegreen}{Problem Solver 1}, \textcolor{orange}{Problem Solver 2}, and \textcolor{red}{Evaluator}.
}
\label{fig:data_sample_complete}

\end{document}